\documentclass{article}

\usepackage[sglblindworkshop, final]{neurips_2025}
\workshoptitle{AI for Science workshop (NeurIPS 2025)}

\usepackage[utf8]{inputenc} % allow utf-8 input
\usepackage[T1]{fontenc}    % use 8-bit T1 fonts
\usepackage{hyperref}       % hyperlinks
\usepackage{url}            % simple URL typesetting
\usepackage{booktabs}       % professional-quality tables
\usepackage{amsfonts}       % blackboard math symbols
\usepackage{nicefrac}       % compact symbols for 1/2, etc.
\usepackage{microtype}      % microtypography
\usepackage{xcolor}         % colors
\usepackage{amsmath}
\usepackage{amssymb}
\usepackage{mathtools}
\usepackage{booktabs}
\usepackage{colortbl}
\usepackage{color, xcolor}
\usepackage{bm}
\usepackage{graphicx}
\usepackage[linesnumbered,ruled,vlined]{algorithm2e}
\usepackage[table]{xcolor}
\usepackage{tcolorbox}
\usepackage{wrapfig}

\title{Sparse Mixture-of-Experts for Multi-Channel Imaging: Are All Channel Interactions Required?}
\newcommand{\affmark}[1]{\textsuperscript{#1}} % superscripts for affiliations
\newcommand{\affil}[2]{\textsuperscript{#1}\,#2} % numbered affiliation lines

\author{%
  Sukwon Yun\affmark{1,2}\textsuperscript{$\dagger$} \quad
  Heming Yao\affmark{3} \quad
  Burkhard Hoeckendorf\affmark{3} \quad
  David Richmond\affmark{3} \\
  \textbf{Aviv Regev}\affmark{2}\thanks{Correspondence to: \texttt{\{regev.aviv,littman.russell\}@gene.com}.} \quad
  \textbf{Russell Littman}\affmark{2}\footnotemark[1] \\
  \\
  \affil{1}{University of North Carolina at Chapel Hill} \\
  \affil{2}{Research and Early Development (gRED), Genentech} \\
  \affil{3}{Biology Research — AI Development (BRAID), Genentech} \\
  \textsuperscript{$\dagger$}\,Work completed during an internship at Genentech.
}

\usepackage{fancyhdr}

\begin{document}

\maketitle

\vspace{-5mm}
\begin{abstract}
Vision Transformers (ViTs) have become the backbone of vision foundation models, yet their optimization for multi-channel domains—such as cell painting or satellite imagery—remains underexplored. A key challenge in these domains is capturing interactions between channels, as each channel carries different information. While existing works have shown efficacy by treating each channel independently during tokenization, this approach naturally introduces a major computational bottleneck in the attention block: channel-wise comparisons leads to a quadratic growth in attention, resulting in excessive FLOPs and high training cost. In this work, we shift focus from efficacy to the overlooked \textit{efficiency challenge} in cross-channel attention and ask: ``Is it necessary to model all channel interactions?". Inspired by the philosophy of Sparse Mixture-of-Experts (MoE), we propose MoE-ViT, a Mixture-of-Experts architecture for multi-channel images in ViTs, which treats each channel as an expert and employs a lightweight router to select only the most relevant experts per patch for attention. Proof-of-concept experiments on real-world datasets—JUMP-CP and So2Sat—demonstrate that MoE-ViT achieves substantial efficiency gains without sacrificing, and in some cases enhancing, performance, making it a practical and attractive backbone for multi-channel imaging.
\end{abstract}

\section{Introduction}

\begin{wrapfigure}{r}{0.32\textwidth}
    \vspace{-7mm}
  \centering
  \includegraphics[width=0.8\linewidth]{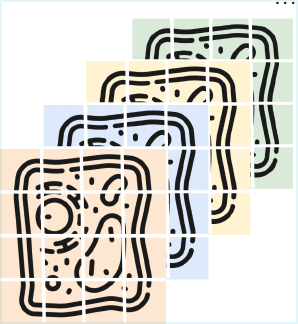}
    \vspace{-3mm}
    \caption{Conceptual illustration of multi-channel imaging. Each patch contains multiple channels represented by different colors.}
    \label{fig:fig_1}
\vspace{-6mm}
\end{wrapfigure}

% 1. ViT and challenges in Multi-channel Image  
Vision Transformers (ViTs) have become the de facto backbone of many modern vision foundation models ~\citep{vision_transformers, dino_v3}. However, their generalization beyond conventional RGB images (i.e., 3-channel inputs) remains underexplored—particularly in the context of multi-channel imaging (Figure~\ref{fig:fig_1}) such as cellular fluorescence microscopy or satellite remote sensing ~\citep{channel_vit}. 
Addressing this gap is important given the growing prevalence of multi-channel datasets in scientific and industrial domains, where the number of spectral or modality-specific channels continues to increase. 
Multi-channel modalities carry distinct and complementary information—each channel may correspond to a unique wavelength, staining protocol, or sensing modality, revealing structures or patterns invisible to other channels ~\citep{channel_vit, JUMP_CP, so2sat}. 
Effectively modeling the relationships across channels is thus crucial for unlocking the full potential of multi-channel imaging ~\citep{channel_vit}.

% 2. Early works and failures
Common approaches to adapt ViTs for multi-channel settings typically adopt a channel-wise design: instead of treating a patch as a multi-channel concatenated token, these models process each channel separately, creating a distinct patch token per channel ~\citep{channel_vit, chada_vit, dicha_vit}. While this design improves downstream performance by better capturing channel-specific interactions, it introduces a critical drawback: the attention cost scales quadratically with both the number of patches ($N$) and channels ($C$), increasing from the standard $\mathcal{O}(N^2)$ to $\mathcal{O}(N^2 C^2)$. This leads to prohibitively high FLOPs and memory usage, severely limiting real-world deployment—especially when $C$ or $N$ are large.

% 3. Our reserach question and answer
In this work, we shift focus from maximizing efficacy to tackling the \textbf{efficiency} directly, raising a natural question:

\begin{tcolorbox}[before skip=0.2cm, after skip=0.2cm, boxsep=0.0cm, middle=0.1cm, top=0.1cm, bottom=0.1cm]
\textit{\textbf{(Q)}} \textit{Do all channels need to participate in attention for every patch? Or can we selectively involve only the most relevant channel interactions per patch?}
\end{tcolorbox}

Our answer is: \emph{No}—not all channel interactions are necessary; \emph{Yes}—we can selectively involve only the relevant ones. Motivated by the Sparse Mixture-of-Experts (MoE) design, where a router activates only the top-$k$ relevant experts for each input, we map channels onto experts and introduce a lightweight \textbf{channel router} that dynamically selects only the most relevant channels for each patch's attention. This reduces complexity from $\mathcal{O}(N^2 C^2)$ to $\mathcal{O}(N^2 C k)$, where $k \ll C$ (often 1–2), dramatically lowering computation while preserving the most important cross-channel interactions.

% 4. Proof-of-concep results
As a proof-of-concept, we evaluate our method on two multi-channel imaging datasets: the cellular perturbation dataset JUMP-CP~\citep{JUMP_CP} (8 channels) and the remote sensing dataset So2Sat~\citep{so2sat} (18 channels).
Using the same backbone and architecture as recent state-of-the-art channel-wise ViTs ~\citep{vision_transformers, chada_vit}, we simply replace the standard multi-head attention in each Transformer layer with our proposed channel-router-based attention. This minimal change allows our model to match—and in some cases outperform—strong baselines, while requiring fewer FLOPs.

% 5. Our contribution recap
By introducing sparsity into the cross-channel dimension, our approach achieves a principled balance between accuracy and efficiency, enabling scalable, cost-efficient adoption of multi-channel ViTs in large-scale and resource-constrained applications.

\vspace{-3mm}

\section{Related Works}
% 1. ViT in biological image: ViT, Channel-ViT, ChAda-ViT, DiChaViT
\textbf{Vision Transformers (ViTs) for Multi-Channel Imaging.} 
% While ViTs have achieved remarkable success on RGB images, their extension to multi-channel imaging data—where each channel often carries distinct and complementary information—has been less explored.  
% To address this gap, pioneering works such as Channel-ViT~\citep{channel_vit} treat each channel separately at the patch level, which improves classification accuracy but overlooks computational cost. ChAda-ViT~\citep{chada_vit} demonstrates the effectiveness of self-supervised learning pretraining equipped with token padding and masking strategies, while DiChaViT~\citep{dicha_vit} enhances feature diversity both across and within channels, improving the richness of learned latent representations.  
% Despite their effectiveness, these approaches still face practical deployment challenges due to the heavy cost of exhaustive cross-channel attention. Although some works~\citep{channel_vit, dicha_vit} adopt hierarchical channel sampling (HCS) where random channels are selected during training, due to randomness we can not guarantee selected channels are relevant to the patch and also when it comes to inference phase, interacting with all the available channels are inevitable. 
% In this work, we address this limitation by introducing a lightweight, channel router that selectively activates only the most relevant channels, enabling efficient yet informative cross-channel interactions.
While ViTs have achieved remarkable success on RGB images, their extension to multi-channel imaging—where each channel often carries distinct and complementary information—remains underexplored.
Channel-ViT~\citep{channel_vit} processes channels separately at the patch level, improving accuracy but increasing computation. ChAda-ViT~\citep{chada_vit} leverages self-supervised pretraining with token padding/masking, while DiChaViT~\citep{dicha_vit} enhances both intra- and inter-channel feature diversity.
However, these methods still incur the high cost of exhaustive cross-channel attention. Although techniques such as Hierarchical Channel Sampling (HCS)~\citep{channel_vit, dicha_vit} reduce cost by randomly selecting channels during training, this randomness can overlook patch-relevant channels and still requires full-channel interaction at inference. We address these limitations with a lightweight channel router that selectively activates only the most relevant channels, enabling efficient yet informative cross-channel interactions.

% 2. Mixture-of-Experts: MoE, GShared, Swith Transformer
\textbf{Sparse Mixture-of-Experts (MoE).  }
The SMoE model builds on the traditional Mixture-of-Experts framework by introducing sparsity to enhance both computational efficiency and model scalability~\citep{shazeer2017outrageously, chen1999improved}. Unlike dense models, SMoE activates only a subset of relevant experts for each task using top-$k$ operation, reducing computational load and enabling effective handling of complex, high-dimensional data. This selective activation has proven advantageous across diverse domains, including vision and language tasks, where it can dynamically adapt different parts of the network to specialized sub-tasks or data types~\citep{vision_moe, gshard, yun2024flexmoemodelingarbitrarymodality, llamamoe2}. 

In this work, we pioneer the adaptation of SMoE to the multi-channel image domain—a setting that remains underexplored yet holds significant potential—leveraging its computational efficiency to model diverse channel interactions in a more lightweight manner, making it readily deployable in practical scenarios.

\begin{figure*}[!th]
    \centering
    \includegraphics[width=\linewidth]{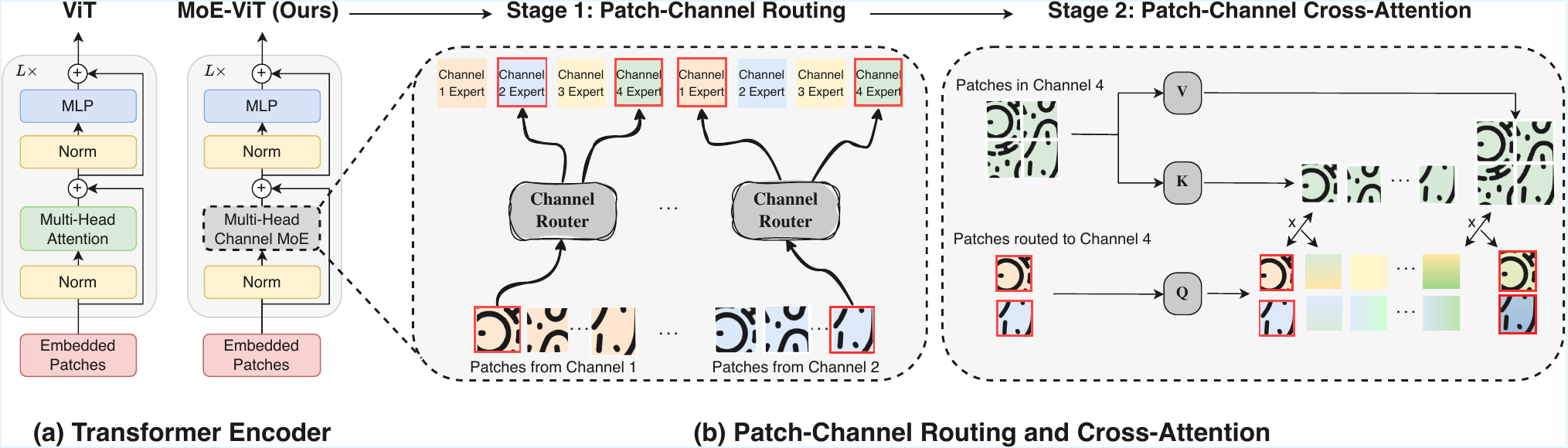}
    \vspace{-5mm}
    \caption{Overview of MoE-ViT. \textbf{(a)} While the vanilla ViT adopts a standard Multi-Head Attention module in the transformer encoder block, MoE-ViT replaces it with a specialized Multi-Head Channel MoE module tailored for multi-channel imaging. \textbf{(b)} Inside the Multi-Head Channel MoE module, input patches pass through the channel router, which allocates each patch to the most relevant channels for interaction (top-2 in this example). Once routing is completed, based on patch and expert indices, we loop each channel to perform patch–channel cross-attention. Specifically, patches routed to a given channel (Channel~4 in this example) serve as the source matrix (Query), while patches belonging to that channel serve as the target matrix (Key and Value) for the cross-attention operation.}    
    \vspace{-5mm}
    \label{fig:framework}
\end{figure*}

% \begin{figure*}[!th]
%     \centering
%     \includegraphics[width=\linewidth]{figs/fig_test.pdf}
%     \vspace{-3mm}
%     \caption{The overview of our framework. Three main innovative components are introduced in our frameworks: an efficient embedding process that is permutation-invariant; a module that captures genetic pathways; and a training framework for cell information aggregation.}
%     \vspace{-6mm}
%     % \label{fig:framework}
% \end{figure*}

\section{Methodology}
\subsection{SMoE Notation}
The SMoE comprises several experts, denoted as $f_{1}, f_{2}, \dots, f_{E}$, where $E$ represents the total number of experts, and a router, $\mathcal{R}$, which selects necessary experts in a sparse fashion. For a given embedding $\mathbf{x}$, the top-$k$ experts are engaged by $\mathcal{R}$ based on the highest scores $\mathcal{R}(\mathbf{x})_j$, where $j$ indicates the expert index. This procedure is formulated as follows:

\begin{equation}
\begin{split}
    \mathbf{y} &= \sum_{j=1}^{k}\mathcal{R}(\mathbf{x})_j\cdot f_j(\mathbf{x}), \\
    \mathcal{R}(\mathbf{x}) &= \text{Top-K}(\text{softmax}(g(\mathbf{x})), k), \\ 
    \mathbf{v} &= \text{softmax}(g(\mathbf{x})), \\
    \text{Top-K}(\mathbf{v}, k) &= 
    \begin{cases}
    \mathbf{v}, & \text{if }\mathbf{v}\text{ is in the top } k, \\
    0, & \text{otherwise}.
    \end{cases}
\end{split}
\label{eq:1}
\end{equation}

where $\mathbf{y}$ denotes the final output of the SMoE layer, computed as the weighted sum of expert outputs $f_j(\mathbf{x})$ scaled by their router-assigned weights $\mathcal{R}(\mathbf{x})_j$. The function $g$ is a trainable module—typically a small Feedforward Neural Network (FFN) with one or a few layers~\citep{slnn_moe, vision_moe}. The $\text{Top-K}(\cdot)$ operation preserves an element of a vector $\mathbf{v}$ only if its probability ranks within the top $K$; otherwise, it is replaced with zero.

\subsection{Our Approach: MoE-ViT}
Motivated by SMoE's principle of selecting only the most relevant experts, we extend this philosophy to multi-channel imaging by treating each channel as an expert. This aligns naturally with the SMoE paradigm, where each expert specializes in capturing a certain context—in our case, channel-specific information—and is activated sparsely so that only the most significant channels are involved. To implement this idea, we replace the Multi-Head Attention module in the ViT architecture with our proposed Channel MoE, in which a channel router (a single-layer FFN) selects the channels to interact with for a given patch. We then perform patch–channel cross-attention: patches routed to a given channel serve as queries, while patches within that channel act as keys and values to capture cross-channel interactions at the patch level. This lightweight design significantly reduces the computational cost compared to prior approaches that required interactions with all available channels. An overview of MoE-ViT is shown in Figure~\ref{fig:framework}. We next describe two main stages: (1) Patch–Channel Routing (Section~\ref{sec:stage_1}) and (2) Patch–Channel Cross-Attention (Section~\ref{sec:stage_2}).

\subsubsection{Patch-Channel Routing \label{sec:stage_1}}
We consider a multi-channel image input $X \in \mathbb{R}^{B \times H \times W \times C}$, where $B$ is the batch size, $(H, W)$ are the image resolutions, and $C$ is the number of channels. Similar to ViT~\citep{vit}, we first partition the image into non-overlapping patches of size $(P, P)$ along the spatial dimensions, resulting in $N = \frac{HW}{P^2}$ patches per channel. Each patch is then flattened into a vector, yielding a sequence $\mathbf{x}_p \in \mathbb{R}^{N \times (P^2 \cdot C)}$, where $N$ is the number of spatial patches and $C$ is the number of channels. We then apply a trainable linear projection to map each vector to a $D$-dimensional token embedding.

In our setting, we treat each channel separately, similar to channel-wise ViTs, and denote the token at spatial position $i$ and channel $j$ as $P_{i,j} \in \mathbb{R}^D$. We add both learnable positional embeddings (for spatial location) and channel embeddings (for channel identity), followed by a feed-forward transformation applied to each token:

\begin{align}
P'_{i,j} &= P_{i,j} + \mathrm{pos}_i + \mathrm{chan}_j,\\
h_{i,j} &= \mathrm{FFN}(P'_{i,j}) \in \mathbb{R}^{D}.
\end{align}

With the token embeddings prepared, existing approaches typically perform multi-head attention over all tokens across all channels, incurring a computational cost of $\mathcal{O}(N^2C^2)$. Instead, we adopt the SMoE philosophy to achieve sparsity in channel interactions. Specifically, we adapt the original SMoE formulation (Equation~\ref{eq:1}) by setting the number of experts equal to the number of channels, assigning one expert per channel so that each expert specializes in channel-specific information.

Formally, we define a set of $C$ \textbf{channel experts} $\{f_k\}_{k=1}^C$, one per channel. Each token (i.e., patch embedding) is routed to a subset of these channel experts for interaction. To this end, we introduce a gating function $g: \mathbb{R}^D \to \mathbb{R}^C$, implemented as a single linear layer in MoE-ViT, which produces routing scores $\mathcal{R}(\cdot)$. At the same time, we obtain the set of top-$k$ selected experts $\mathcal{E}(\cdot)$ as follows:

\begin{align}
\mathcal{R}(h_{i,j}) &= \text{Top-K}(\text{softmax}(g(h_{i,j})), k) \in\mathbb{R}^{C} \\
\mathcal{E}(i,j) &= \{\, c \mid \mathcal{R}(h_{i,j})[c] \neq 0 \,\}
\end{align}

\noindent where $\mathcal{R}(h_{i,j})$ denotes the routing scores\footnote{For balancing, following standard MoE regularization techniques (e.g., importance/load losses~\citep{shazeer2017outrageously}), we employ this regularizer to encourage uniform expert activation and mitigate routing collapse.} of the patch at $(i,j)$, and $\mathcal{E}(i,j)$ denotes the set of expert (channel) indices selected for that patch. By doing so, each patch is sparsely routed to $k$ channels and is now ready for patch–channel cross-attention.

% The router implements a sparse mixture-of-experts (MoE) over channels. For each token $h_{i,j}$, $r_{i,j}$ scores its affinity to every expert $E_k$. The hard set $\mathcal{E}(i,j)$ determines which experts will process the token; this can be viewed as a \emph{dispatch} operator. During training, the dispatch can be implemented as hard top-$K$ (for compute sparsity) with optional soft weighting inside the selected set using the normalized scores $r_{i,j,k}$. In our default formulation we keep uniform weighting at aggregation time (Eq.~\eqref{eq:agg}), but a score-weighted variant is a straightforward extension.

\subsubsection{Patch-Channel Cross-Attention \label{sec:stage_2}}
Once routing is complete, next is to capture interactions between each patch and its selected channels (i.e., the patches within those channels). Specifically, for each channel $k$, we form two matrices:
(i) a \textit{source} matrix $S_k$, containing all patches routed to channel $k$; and
(ii) a \textit{target} matrix $T_k$, containing all patches from channel $k$, as illustrated in Figure~\ref{fig:framework} (b). Formally, these matrices are defined as:

\begin{align}
S_k &= \big\{\, h_{i,j}\ \big|\ k\in\mathcal{E}(i,j) , \forall i,j \, \big\}
\ \in \ \mathbb{R}^{N_k \times D},\\
T_k &= \big\{\, h_{i,k}\ \big|\ \forall i \,\big\}
\ \in \ \mathbb{R}^{M_k \times D}.
\end{align}

where $N_k$ is the number of patches routed to channel $k$, and $M_k$ is the number of patches in channel $k$. We perform cross-attention from the sources to the targets, employing a shared projection for the sources and channel-specific projections for the targets, using scaled dot-product attention as follows:

\begin{align}
Q_k &= S_k W^Q,\qquad
K_k \ =\ T_k W^K_k,\qquad
V_k \ =\ T_k W^V_k, \\
A_k &= \mathrm{softmax}\!\left(\frac{Q_k K_k^\top}{\sqrt{d_k}}\right) \in \ \mathbb{R}^{N_k \times M_k}, \qquad
O_k \ =\ A_k V_k \ \in \ \mathbb{R}^{N_k \times D}.
\end{align}

Here, $W^Q$, $W^K_{\cdot}$, and $W^V_{\cdot} \in \mathbb{R}^{D \times D}$ are learnable linear projections. It is worth noting that the channel-specific key ($W^K_k$) and value ($W^V_k$) projections serve the role of channel-specific experts in the sparse MoE formulation (Equation~\ref{eq:1}), with each expert $f_k$ specialized for processing information from its corresponding channel.

Finally, for each original token $(i,j)$ we aggregate the outputs from all experts it visited. By default we use uniform averaging over the selected experts:

\begin{equation}
\label{eq:agg}
\hat{h}_{i,j} = \sum\limits_{k \in \mathcal{E}(i,j)} \mathcal{R}(h_{i,j})_k \cdot O_k[\text{idx}(i,j)]
\end{equation}

where $\hat{h}_{i,j}$ is the updated embedding of patch $(i,j)$, and $\text{idx}(i,j)$ retrieves the row of $O_k$ corresponding to $(i,j)$ in $S_k$. In summary, keeping the backbone and all other modules unchanged, but replacing the standard Multi-Head Attention in the Transformer encoder with our proposed MoE-ViT channel-routing mechanism, allows each patch to be updated only through interactions with its relevant channels, thereby reducing computational cost. The overall procedure of MoE-ViT is summarized in Algorithm~\ref{alg:alg1}.

\begin{algorithm}[t]\label{alg:alg1}
\caption{Patch-Channel Routing and Cross-Attention}
\KwIn{Patches $P_{i,j} \in \mathbb{R}^{D}$ for position $i$ and channel $j$}
\KwOut{Updated patch embeddings $\hat{h}_{i,j}  \in \mathbb{R}^{D}$}

{\textcolor{blue}{/* Token Embedding + Positional \& Channel Embedding */}}
\For{$j = 1, \dots, C$}{
    \For{$i = 1, \dots, H \times W$}{
        $P'_{i,j} \leftarrow P_{i,j} + \text{pos}_i + \text{chan}_j$\;
        $h_{i,j} \leftarrow \text{FFN}(P'_{i,j})$ \tcp*[r]{Shape: $[D]$}
    }
}
\BlankLine

{\textcolor{blue}{/* Patch-Channel Routing */}}
\ForEach{$h_{i,j}$}{
    $\mathcal{R}(h_{i,j}) \leftarrow \text{Top-K}(\text{softmax}(g(h_{i,j})), k)$\;
    $\mathcal{E}(i,j) \leftarrow \{\, c \mid \mathcal{R}(h_{i,j})[c] \neq 0 \,\}$\;
}
\BlankLine

{\textcolor{blue}{/* Patch-Channel Cross-Attention */}}
\For{$k = 1, \dots, K$}{
    $S_k \leftarrow \{\, h_{i,j} \mid k\!\in\!\mathcal{E}(i,j), \forall i,j \,\}$ \tcp*[r]{Shape: $[N_k \times D]$}
    $T_k \leftarrow \{\, h_{i,k} \mid \forall i \,\}$ \tcp*[r]{Shape: $[M_k \times D]$}
    $Q_k \leftarrow S_k W^Q$\;
    $K_k \leftarrow T_k W^K_k$\;
    $V_k \leftarrow T_k W^V_k$\;
    $A_k \leftarrow \text{softmax}\!\left( \frac{Q_k K_k^T}{\sqrt{d_k}} \right)$ \tcp*[r]{Shape: $[N_k \times M_k]$}
    $O_k \leftarrow A_k V_k$ \tcp*[r]{Shape: $[N_k \times D]$}
}
%\BlankLine

{\textcolor{blue}{/* Aggregation */}}
\ForEach{$(i,j)$}{
    $\hat{h}_{i,j} \leftarrow \sum\limits_{k \in \mathcal{E}(i,j)} \mathcal{R}(h_{i,j})_k \cdot O_k[\text{idx}(i,j)]$ \tcp*[r]{Shape: $[D]$}
}
\end{algorithm}

\paragraph{Complexity.} For the \textbf{attention term}, in standard multi-head self-attention over multi-channel inputs, each of the $NC$ tokens ($N$ spatial patches per channel and $C$ channels) attends to all others, yielding $\mathcal{O}(BN^2 C^2D)$ for $(QK^\top, AV)$. In our MoE design, the router assigns each token to only $k \ll C$ channels. Let $N_k$ and $M_k$ be the numbers of source and target tokens for channel $k$. With $M_k \approx N$ and $\sum_{k=1}^C N_k = k N C$ (uniform routing), the total becomes $\mathcal{O}(BN^2 C k D)$,
a $\tfrac{k}{C}$ fraction of the full attention cost.

For the \textbf{projection term}, standard attention applies a shared $W^Q$ to all tokens and a shared $W^K, W^V$ per channel, costing $\mathcal{O}(BN C D^2)$ per type, i.e., $3 \times \mathcal{O}(BN C D^2)$. Our design keeps $W^Q$ shared but applies $W^K_k$ and $W^V_k$ only to $k$ selected channels, giving $\mathcal{O}(BN k D^2)$
per type. 

The attention and projection costs equal the standard case only when $k = C$ and scales down with smaller $k$. Since attention scales as $\mathcal{O}(N^2)$ and projection as $\mathcal{O}(N)$, reducing the attention cost dominates overall FLOPs savings for $N \gg D$, enabling substantial efficiency gains while preserving cross-channel modeling.

\section{Experiments}

\noindent \textbf{Datasets}
We evaluate MoE-ViT and baselines across two datasets in different domains (microscopy and satellite imagery), showing its general utility in reducing compute (FLOPs) while increasing accuracy. (1) \textbf{JUMP-CP}\citep{JUMP_CP} is a microscopy dataset with images of cell crops (224×224) that underwent chemical perturbations. We focus on plate BR00116991 as in~\citep{channel_vit}, which contains 127k training, 45k validation, and 45k test images. The images have 5 fluorescence and 3 bright field channels, totaling 8 channels. The task is to classify treatment groups from 161 classes. (2) \textbf{So2Sat}~\citep{so2sat} contains satellite imagery from two remote sensors with 8 channels from Sentinel-1 and 10 channels from Sentinel-2, totaling 18 channels. Each crop is 32×32. The task is to classify 17 climate zones. We use 352k training and 24k test images.

\noindent \textbf{Baselines} As MoE-ViT builds upon channel-wise ViTs by introducing a simple yet effective modification to the multi-head attention (MHA) module in the Transformer encoder, we compare it against the most representative and well-established ViT-based baselines. Specifically, we consider a standard ViT (ViT-Small)\citep{vit} that integrates all channels jointly, and two channel-wise variants: ChAda-ViT\citep{chada_vit} and DiChaViT~\citep{dicha_vit}. ChAda-ViT follows the Channel-ViT~\citep{channel_vit} design and incorporates enhancements originally intended for self-supervised training, while DiChaViT further improves intra- and inter-channel representations in the latent space. For fair comparison, we trained ChAda-ViT from scratch in a supervised setting, without self-supervised pretraining.

% \noindent \textbf{Experimental Settings}
% We utilized the experimental set up defined by Channel-ViT ~\citep{channel_vit}. We trained MoE-ViT as well as baselines vanilla ViT ~\citep{vision_transformers}, ChAda-ViT ~\citep{chada_vit} and DiChaViT ~\citep{dicha_vit} for 100 epochs on Jump-CP and So2Sat evaluating the representation of the CLS token on the top-1 classification accuracy. We note, that due to restrictive licensing we were unable to compare against the original Channel-ViT model. We use GFLOPs as a proxy for compute of the attention mechanism for each model. 

\textbf{Experimental Settings} We follow the experimental setup of Channel-ViT~\citep{channel_vit}, training MoE-ViT and baselines—vanilla ViT~\citep{vision_transformers}, ChAda-ViT~\citep{chada_vit}, and DiChaViT~\citep{dicha_vit}—for 100 epochs on JUMP-CP and So2Sat, evaluating top-1 classification accuracy using the CLS token representation. Due to licensing restrictions, Channel-ViT results are unavailable. GFLOPs are reported as a proxy for attention cost. Baselines follow their original configurations, while MoE-ViT adapts ChAda-ViT with our sparse MoE attention. All models use ViT-Small backbones with AdamW~\citep{adam_w} and hierarchical channel sampling (HCS)~\citep{channel_vit} during training. We employ patch size 16 and 8 for JUMP-CP and So2Sat, respectively.

\begin{table}[!h]
\centering
\caption{Performance comparison on the JUMP-CP dataset (ViT-S/16 as the backbone). \textbf{Attn.} denotes attention, and \textbf{\# of Act. Param.} denotes the number of activated parameters in the entire model. Complexity and GFLOPs are measured during inference. For MoE-ViT, calculations include the channel router and channel-specific key/value projections in cross-attention, while keeping all other Transformer encoder modules identical to the baseline.}
\vspace{-2mm}
\resizebox{0.95\textwidth}{!}{
\begin{tabular}{@{}lcccc@{}}
\toprule
\textbf{} & \multicolumn{4}{c}{\textbf{JUMP-CP}} \\ \cmidrule(lr){2-5}
 & \textbf{Acc} & \textbf{Attn. Complexity} & \textbf{Attn. GFLOPs} & \textbf{\# of Act. Param.} \\
\midrule
ViT & 56.24 & $\mathcal{O}(N^2d)$ & 0.29G & 22.28M  \\
ChAda-ViT & 68.16 & $\mathcal{O}(N^2C^2d)$ & 5.65G & 21.60M  \\
DiChaViT & 68.49 & $\mathcal{O}(N^2C^2d)$ & 5.65G & 21.60M  \\ \bottomrule
\rowcolor{gray!15} \textbf{MoE-ViT (Top-k=$1$)} & 64.06 & $\mathcal{O}(N^2Ckd)$ & 2.33G & 21.62M  \\
\rowcolor{gray!15} \textbf{MoE-ViT (Top-k=$2$)} & 66.44 & $\mathcal{O}(N^2Ckd)$ & 2.81G & 25.18M  \\
\rowcolor{gray!15} \textbf{MoE-ViT (Top-k=$C$)} & 70.16 & $\mathcal{O}(N^2C^2d)$ & 5.65G & 46.47M  \\
\bottomrule
\end{tabular}
}
\label{tab:jump_cp_results}
\end{table}

\begin{table}[!h]
\centering
\caption{Performance comparison in So2Sat dataset (ViT-S/8 as the backbone). \textbf{Attn.} denotes attention, and \textbf{\# of Act. Param.} denotes the number of activated parameters in the entire model. Complexity and GFLOPs are measured during inference. For MoE-ViT, calculations include the channel router and channel-specific key/value projections in cross-attention, while keeping all other Transformer encoder modules identical to the baseline.}
\vspace{-2mm}
\resizebox{0.95\textwidth}{!}{
\begin{tabular}{@{}lcccc@{}}
\toprule
\textbf{} & \multicolumn{4}{c}{\textbf{So2Sat}} \\ \cmidrule(lr){2-5}
 & \textbf{Acc} & \textbf{Attn. Complexity} & \textbf{Attn. GFLOPs} & \textbf{\# of Act. Param.} \\
\midrule
ViT & 58.93 & $\mathcal{O}(N^2d)$ & 0.02G & 21.76M  \\
ChAda-ViT & 63.94 & $\mathcal{O}(N^2C^2d)$ & 0.47G & 21.35M  \\
DiChaViT & 63.80 & $\mathcal{O}(N^2C^2d)$ & 0.47G & 21.35M  \\ \bottomrule
\rowcolor{gray!15} \textbf{MoE-ViT (Top-k=$1$)} & 63.12 & $\mathcal{O}(N^2Ckd)$ & 0.35G & 21.43M  \\
\rowcolor{gray!15} \textbf{MoE-ViT (Top-k=$2$)} & 64.66 & $\mathcal{O}(N^2Ckd)$ & 0.36G & 24.98M \\
\bottomrule
\end{tabular}
}
\label{tab:so2sat_results}
\end{table}

\subsection{Primary Results}
Tables~\ref{tab:jump_cp_results} and~\ref{tab:so2sat_results} summarize the performance and efficiency trade-offs of MoE-ViT compared to existing baselines. We make the following key observations: 

\textbf{(1) Balancing accuracy and efficiency.} On both datasets, the vanilla ViT achieves the lowest attention cost (0.29G GFLOPs in JUMP-CP and 0.02G GFLOPs in So2Sat) but also the lowest top-1 classification accuracy (56.24\% and 58.93\%, respectively), reflecting the limitations of embedding all channels jointly without explicitly modeling cross-channel interactions. In contrast, channel-wise ViTs such as ChAda-ViT and DiChaViT substantially improve accuracy (+12.25\% on JUMP-CP and +5.01\% on So2Sat) but incur a significant increase in attention cost (5.65G GFLOPs in JUMP-CP and 0.47G GFLOPs in So2Sat). Our proposed MoE-ViT strikes a favorable balance between these extremes: with Top-$k=2$, it reduces attention cost by roughly 50\% (2.81G GFLOPs in JUMP-CP) compared to ChAda-ViT while sacrificing only 2.05\% accuracy; notably, on So2Sat, it even improves accuracy by +0.72\% while using 23\% fewer GFLOPs, demonstrating competitive performance at markedly lower computational cost.

\textbf{(2) Full-channel MoE-ViT (Top-$k = C$).} When the number of active experts is set equal to the total number of channels, MoE-ViT matches the attention cost of cross-channel attention baselines (5.65G GFLOPs in JUMP-CP and 0.47G GFLOPs in So2Sat) but achieves higher accuracy---a gain of +1.67\% on JUMP-CP and +0.72\% on So2Sat over the best baseline. This improvement is likely due to the use of channel-specific key and value projections in MoE-ViT. By assigning each channel its own expert, the model effectively increases its representational capacity. Despite this increase in parameters, the theoretical computational complexity is not elevated, since the MoE formulation only activates a subset of experts per forward pass.

\textbf{(3) Extreme sparsity (Top-$k = 1$).} Under the most restrictive configuration, MoE-ViT achieves the lowest attention cost among all non-trivial methods (2.33G GFLOPs in JUMP-CP and 0.35G GFLOPs in So2Sat) while still outperforming vanilla ViT, demonstrating robustness to high sparsity in expert activation.

\textbf{(4) Dataset dependency.} The relative efficiency gains of MoE-ViT are more pronounced in JUMP-CP, which contains larger images (224$\times$224), than in So2Sat (32$\times$32). This suggests that the computational savings from sparse expert routing scale with the number of spatial tokens, and datasets with larger spatial resolutions stand to benefit more from the proposed method.

\textbf{(5) Activated parameters and model capacity.} Interestingly, vanilla ViT has a higher number of activated parameters than MoE-ViT with Top-$k=1$ (22.28M vs. 21.62M in JUMP-CP), despite its lower attention complexity and inferior accuracy. This is because it concatenates all channels into a single vector, resulting in a higher-dimensional representation that is projected into the embedding space. Moreover, MoE-ViT with larger Top-$k$ values can involve significantly more parameters during inference (e.g., 46.47M at Top-$k=C$ in JUMP-CP) without incurring a quadratic increase in FLOPs. This ability to leverage substantially greater model capacity while maintaining modest computational cost is a key advantage—and arguably the most compelling aspect—of the MoE design.

Overall, these results demonstrate that MoE-ViT not only matches or surpasses the accuracy of cross-channel attention baselines, but also offers fine-grained control over the efficiency--accuracy trade-off via the Top-$k$ parameter, making it adaptable to diverse computational budgets and dataset characteristics.

\begin{figure*}[!ht]
    \centering
    \includegraphics[width=\linewidth]{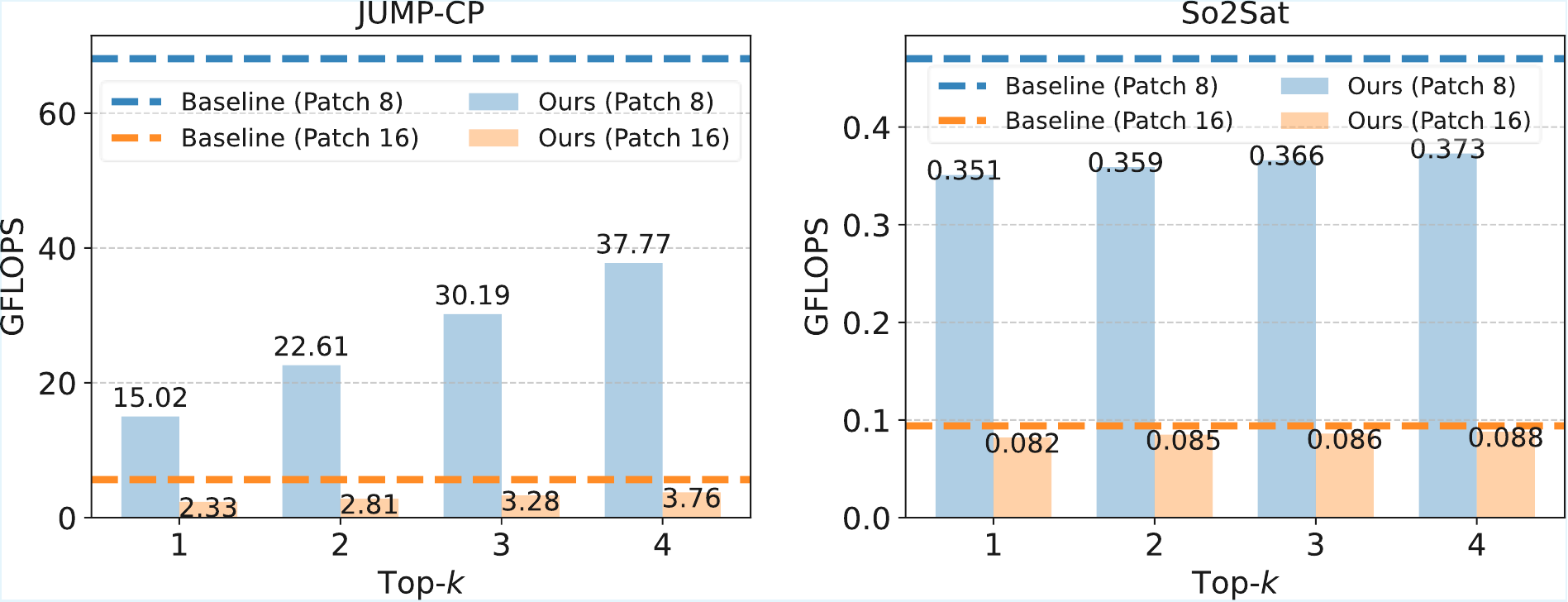}
    \vspace{-5mm}
    \caption{Ablation study comparing the Baseline (ChAda-ViT) and Ours (MoE-ViT) for varying top-$k$ values (i.e., channels) with patch sizes of 8 and 16 on both the JUMP-CP and So2Sat datasets.}    
    \vspace{-5mm}
    \label{fig:ablation}
\end{figure*}

\subsection{Ablation Study}

To better understand the efficiency gains from the top-$k$ design in MoE-ViT, we conducted an ablation study varying both the patch size and the number of activated experts (top-$k$). Figure~\ref{fig:ablation} illustrates the GFLOPs of the attention mechanism under these configurations for both the JUMP-CP and So2Sat datasets. We make the following observations: \textbf{(1)} Across both datasets, decreasing the patch size from 16 to 8 drastically increases the computational cost of attention—from 2.81 to 22.61 GFLOPs (8×) in JUMP-CP and from 0.085 to 0.359 GFLOPs (4×) in So2Sat at top-$k$ = 2—as reducing the patch dimension ($P$) naturally increases the number of tokens ($N = \frac{HW}{P^2}$). \textbf{(2)} With a fixed patch size, increasing top-$k$ from 1 to 4 gradually increases GFLOPs (e.g., from 15.02 to 22.61, 30.19, and 37.77 in JUMP-CP), since more experts are involved in interactions per patch. \textbf{(3)} Despite this increase, MoE-ViT’s GFLOPs remain well below those of the baseline (ChAda-ViT), which exceeds 65 GFLOPs due to attending to all available channels. This highlights the effectiveness of adopting a top-$k$ design to control sparsity.

% 3. Scaling Up (# of Channels, Patch Size, MLP layer as SMoE)

\section{Future Directions}
In this work, we presented a proof-of-concept demonstrating how a sparse mixture-of-experts (MoE) formulation can substantially reduce the FLOPs of channel-wise ViT architectures without sacrificing accuracy. Building on this work, we outline three main directions for future research.

\noindent \textbf{Training and Inference Time Optimization. } 
While FLOPs provide a proxy for efficiency, actual training and inference time also depends on hardware utilization. To fully realize the benefits of MoE architectures, future work could explore tighter co-optimization between architecture and hardware. Leveraging specialized MoE libraries such as DeepSeek-MoE~\citep{deepseek_moe} or FastMoE~\citep{fast_moe}—which provide optimized kernels and routing strategies for modern accelerators—could yield further gains.

% In practice, training and inference time is dependent on a number of things in addition to FLOPS, namely efficient use of accelerators; To reduce training and inference time, the benefits of mixture of experts architectures comes at the optimization between architecture and hardware. Future work could benefit from leveraging specialized MoE libraries such as DeepSeek-MoE ~\citep{deepseek_moe} or FastMoE ~\citep{fast_moe}, which provide optimized kernels and routing strategies designed to maximize efficiency on modern accelerators.

\noindent \textbf{Increasing Sparsity in the Target Matrix. } Our current formulation reduces computation by assigning channel-specific experts, but the framework can be generalized further. For example, one could incorporate a patch-specific router that dynamically allocates experts across spatial regions, thereby further sparsifying the target matrix and reducing computational cost. Patch allocation could be made even sparser by combining our approach with batch-prioritized routing strategies, such as those proposed by Riquelme et al.~\citep{vision_moe}, which prioritizes patches by their routing scores.

\noindent \textbf{Expanding to More Biological Datasets. } Finally, there exists a rich set of multi-channel, large 2D (digital pathology), and 3D datasets both within biology and beyond where the proposed MoE-ViT framework could be applied. 

\section{Conclusion}
% In this work, we addressed a key limitation of channel-wise ViTs: their prohibitive computational complexity in multi-channel imaging domains. By reinterpreting channels as experts and introducing a lightweight channel router, we proposed MoE-ViT, a sparse mixture-of-experts formulation that selectively activates only the most relevant channels per patch. This design reduces the quadratic complexity of cross-channel attention while not sacrificing and sometimes improving—classification accuracy. Through experiments on two distinct datasets, JUMP-CP microscopy and So2Sat satellite imagery, we demonstrated that MoE-ViT consistently achieves substantial efficiency gains in FLOPs while outperforming strong baselines such as ChAda-ViT and DiChaViT. Our ablation studies further revealed that patch size exerts the greatest influence on computational cost, while Top-$k$ provides a tunable axis for balancing capacity and efficiency at relatively low additional cost. Taken together, these results show that sparse expert routing provides a principled way to scale ViTs to high-channel regimes without incurring prohibitive costs.  

In this work, we addressed a key limitation of channel-wise ViTs—their prohibitive computational cost in multi-channel imaging domains. By reinterpreting channels as experts and introducing a lightweight channel router, we proposed MoE-ViT, a sparse mixture-of-experts formulation that selectively activates only the most relevant cross-channel interactions per patch. 
This design reduces the quadratic complexity ($\mathcal{O}(N^2 C^2)$) of cross-channel attention to $\mathcal{O}(N^2 C k)$—linear in $C$ for fixed $k$—while preserving classification accuracy. Experiments on two distinct datasets, JUMP-CP microscopy and So2Sat satellite imagery, show that MoE-ViT consistently achieves substantial FLOPs reductions while outperforming strong baselines such as ChAda-ViT and DiChaViT. Ablation studies further reveal that patch size has the largest impact on computational cost, while Top-$k$ provides a tunable trade-off between capacity and efficiency at relatively low additional cost. Overall, these results demonstrate that sparse expert routing offers a principled approach to scaling ViTs to high-channel regimes without incurring prohibitive computation.

\begin{ack}
H.Y., B.H., D.R., A.R., and R.L. are employees of Roche. H.Y., B.H., D.R. and A.R. have equity in Roche. Work completed during an internship at Genentech (SY). A.R. is a cofounder and equity holder of Celsius Therapeutics, an equity holder in Immunitas and was a scientific advisory board member of ThermoFisher Scientific, Syros Pharmaceuticals, Neogene Therapeutics and Asimov until 31 July 2020. From 1 August 2020, A.R. is an employee of Genentech and has equity in Roche.
\end{ack}

\newpage

\bibliographystyle{plain}
\bibliography{references}
%\bibliographystyle{unsrt}

% \section*{References}

% References follow the acknowledgments in the camera-ready paper. Use unnumbered first-level heading for
% the references. Any choice of citation style is acceptable as long as you ar
% consistent. It is permissible to reduce the font size to \verb+small+ (9 point)
% when listing the references.
% Note that the Reference section does not count towards the page limit.
% \medskip

% {
% \small

% [1] Alexander, J.A.\ \& Mozer, M.C.\ (1995) Template-based algorithms for
% connectionist rule extraction. In G.\ Tesauro, D.S.\ Touretzky and T.K.\ Leen
% (eds.), {\it Advances in Neural Information Processing Systems 7},
% pp.\ 609--616. Cambridge, MA: MIT Press.

% [2] Bower, J.M.\ \& Beeman, D.\ (1995) {\it The Book of GENESIS: Exploring
%   Realistic Neural Models with the GEneral NEural SImulation System.}  New York:
% TELOS/Springer--Verlag.

% [3] Hasselmo, M.E., Schnell, E.\ \& Barkai, E.\ (1995) Dynamics of learning and
% recall at excitatory recurrent synapses and cholinergic modulation in rat
% hippocampal region CA3. {\it Journal of Neuroscience} {\bf 15}(7):5249-5262.
% }

%%%%%%%%%%%%%%%%%%%%%%%%%%%%%%%%%%%%%%%%%%%%%%%%%%%%%%%%%%%%

\appendix

\end{document}